\documentclass[11pt,a4paper]{article}
\usepackage[hyperref]{acl2020}
\usepackage{times}
\usepackage{latexsym}

\usepackage{microtype}
\usepackage{lipsum}
\usepackage{booktabs}
\usepackage{tabularx}

\usepackage{tikz}
\usetikzlibrary{calc}
\usetikzlibrary{positioning,arrows,automata}
\usetikzlibrary{shapes.geometric}
\usepackage{pgfplots}
\usepgfplotslibrary{groupplots}
\usepackage{subcaption}
\usepackage{ulem}
\usepackage{graphicx}
\usetikzlibrary{fit}
\usepackage{cancel}
\usepackage{amsmath}
\aclfinalcopy 

\newcommand\BibTeX{B\textsc{ib}\TeX}

\title{Adapting End-to-End Speech Recognition for Readable Subtitles}

\author{Danni Liu, Jan Niehues, Gerasimos Spanakis \\
  Department of Data Science and Knowledge Engineering, \\
  Maastricht University \\
  \normalsize
  \texttt{\{danni.liu,jan.niehues,jerry.spanakis\}@maastrichtuniversity.nl}}

\begin{document}
\maketitle
\begin{abstract}

Automatic speech recognition (ASR) systems are primarily evaluated on transcription accuracy.
However, in some use cases such as subtitling, verbatim transcription would reduce output readability given limited screen size and reading time.
Therefore, this work focuses on ASR with output compression, a task challenging for supervised approaches due to the scarcity of training data.
We first investigate a cascaded system, where an unsupervised compression model is used to post-edit the transcribed speech. 
We then compare several methods of end-to-end speech recognition under output length constraints. 
The experiments show that with limited data far less than needed for training a model from scratch, we can adapt a Transformer-based ASR model to incorporate both transcription and compression capabilities.
Furthermore, the best performance in terms of WER and ROUGE scores is achieved by explicitly modeling the length constraints within the end-to-end ASR system.
\end{abstract}

\section{Introduction}
Automatic speech recognition (ASR) has become ubiquitous in human interaction with digital devices, such as keyboard voice inputs \cite{he2019streaming} and virtual home assistants \cite{li2017acoustic}.
While transcription accuracy is often the primary goal when designing ASR systems, in some use cases the readability of outputs is crucial to user experience.
A prominent example is subtitling for TV.
In this case, the audience need to multitask, i.e. simultaneously watch video contents, listen to speech utterances, and read subtitles.
To avoid a visual overload, not every spoken word needs to be displayed.
Meanwhile, the shortened subtitles must still retain the meaning of the spoken content.
Moreover, large deviations from the original utterance are also undesirable, as the disagreement with auditory input would create a distraction.

To compress the subtitles, one straightforward approach is to post-process ASR transcriptions.
The task of sentence compression has been well-studied \cite{knight2002summarization, clarke-lapata-2006-models, rush-etal-2015-neural}.
In extractive compression \cite{filippova-etal-2015-sentence, angerbauer2019automatic}, only deletion operations are performed on the input.
Despite the simplicity, this approach tends to produce outputs that are less grammatical \cite{knight2002summarization}. 
On the other hand, abstractive compression \cite{cohn-lapata-2008-sentence, rush-etal-2015-neural, chopra-etal-2016-abstractive, yu-etal-2018-operation} involves more sophisticated input reformulation, such as word reordering and paraphrasing \cite{clarke-lapata-2006-models}.
For the task of compressing subtitles, however, the extent of rewriting must be controlled in order to retain consistency with spoken utterances.

From a practical point of view, building a sentence compression system typically requires training corpora where the target sequences are summarized.
For most languages and domains, there exists scarcely any resource suitable for supervised training.
This low-resource condition is even more severe for audio inputs.
To the best of our knowledge, currently there is no publicly available spoken language compression corpora.

Given the challenges outlined above, this work investigates ASR with output compression. We test our approaches on German TV subtitles. 
The combination of this task and the use case is to the extent of our knowledge previously unexplored.

The first contribution of this work is a comparison of 
cascaded and end-to-end approaches to generating compressed ASR transcriptions, where
the former consists of separate ASR and compression modules, and the latter integrates transcription and compression.
The experiments show that our sentence compression module trained in an unsupervised fashion tends to excessively paraphrase, whereas the end-to-end model can be better adapted to the task of interest.
Secondly, we show that, after fine-tuning on a small adaption corpus, an ASR model can perform transcription and compression simultaneously.
Without being given explicit length constraints, the adapted model shows increased recognition accuracy on rare words as well as paraphrasing capabilities to produce shorter outputs.
Furthermore, by explicitly encoding the length constraints, we achieve further performance gains in addition to those brought by adaptation.

\section{Task}
The task of creating readable subtitles for video contents has several unique properties. 
First, due to limited screen size and reading time, not every spoken word needs to be transcribed, especially when utterances are spoken fast. 
A full transcription could even hamper user experience due to poor readability.
Second, although output shortening is typically realized by deleting non-essential words, the output is not only deletion-based.
A real-life example from the German TV program Tagesschau\footnote{\url{https://www.tagesschau.de/}} shown in Table \ref{tab:example.task} contain rephrasing (from ``freed from'' to ``without'') in addition to word removal (dropping  the word ``ethically'').
A further requirement is that the subtitles should stay reasonably authentic to the spoken contents, only modifying them when necessary.
Otherwise, the disagreement with audio contents could become distracting to the audience.

Within the framework of common NLP tasks, the task of generating readable subtitles combines ASR and abstractive compression, while being subjected to the additional requirements as outlined above.


\begin{table}[h]
	\centering
	\small 
	\begin{tabularx}{\columnwidth}{X}
	\toprule
		\textbf{Spoken}: \underline{Befreit vom} fraktionszwang soll das Parlament wohl nach der Sommerpause die \underline{ethisch} schwierige Frage debattieren.
		{\color{gray} (\underline{Freed from} pressure from the coalition party, the parliamentary should debate the \underline{ethically} difficult question after the summer break.) }\\
		\textbf{Subtitle}: \underline{Ohne} Fraktionszwang soll das Parlament wohl nach der Sommerpause die schwierige Frage debattieren. 
		{\color{gray} (\underline{Without} pressure from the coalition party, the parliamentary should debate the difficult question after the summer break.)} \\
	\bottomrule
	\end{tabularx}	
	\caption{Examples of TV subtitles compared to actual spoken words. Underlined words are the differences between the spoken words and the subtitles.}\label{tab:example.task}
\end{table}


\section{ASR with Output Length Constraints}
\subsection{Baseline ASR Model with Adaptation}
For the baseline ASR model, we use the Transfomer architecture \cite{vaswani2017attention} similar to that by \citet{pham2019very}.
As there is no spoken language compression corpus available to us that is large enough for training an end-to-end model from scratch,
we first train an ASR model without output compression, and then adapt it to our task of interest using a small web-scraped corpus.
In the first training stage, the model is solely trained for transcribing speech.
In the fine-tuning stage, we let the model continue training at a reduced learning rate on the adaptation corpus with shortened transcriptions.
The intended goal of adaptation is to let the model learn the compression task on the basis of the transcription capability acquired before.

\subsection{End-to-End Length-Constrained ASR} \label{subsec:e2e.len}
\begin{figure}[h]
		\tikzstyle{vector}=[rectangle,fill=none,draw=black,text=black,minimum width=0.5cm,minimum height=0.5cm,line width=0.5pt]
		\tikzstyle{io}=[circle,fill=white,draw=black,text=black,minimum width=0.5cm,minimum height=0.5cm]
		\tikzstyle{void}=[rectangle,fill=none,draw=none,text=black,minimum width=0.5cm,minimum height=0.5cm,inner sep=0,outer sep=0]
		\tikzstyle{att}=[->, draw=black]
		\tikzstyle{att1}=[->, draw=gray]
		\tikzstyle{inp}=[->, draw=black]
		\tikzstyle{inpB}=[->,bend left=34, draw=black]
		\tikzstyle{recF}=[->, draw=black]
		\tikzstyle{recB}=[<-, draw=black]
		\centering
		\begin{tikzpicture}[>=stealth',shorten >=1pt,auto,node distance=1cm,
		triangle/.style = {regular polygon, regular polygon sides=3 }]
		
			
			
		
			
			\node[state][vector] at (2,1.5) (s1)              {$\mathbf{y}_0$}; 
			\node[state][vector] (s2) [right of=s1]        {$\mathbf{y}_1$}; 
			\node[state][vector] (s3) [right of=s2]        {$\mathbf{y}_2$}; 
			\node[state][void] (s4) [right of=s3]           {$...$};
			\node[state][vector,inner sep=0.05cm,] (s5) [right of=s4]        {$\mathbf{y}_t$};

				\node[state][void] at (2,-0.5) (te1) {\small   $t$}; 
				\path (te1) edge[att1] (s1); 
				\node[state][void] (te2) [right of=te1]{\small  $t-1$};
				\path (te2) edge[att1] (s2); 
				\node[state][void] (te3) [right of=te2]{ \small  $t-2$}; 
				\path (te3) edge[att1] (s3);
				
				\node[state][void] (te4) [right of=te3]{$...$}; 
				\node[state][void] (te5) [right of=te4]{\small   $0$}; 
				\path (te5) edge[att1] (s5);

			\node[state][io] at (2,3) (y1)                 {$w_1$}; 
			\node[state][io] (y2) [right of=y1]        {$w_2$}; 
			\node[state][io] (y3) [right of=y2]        {$w_3$};
			\node[state][io, fill=red, draw=none] at (6,3) (y5)                 {}; 
			\node[state][rectangle,fill=white,draw=none,minimum width=0.25cm,minimum height=0.25cm,inner sep=0,outer sep=0] at (6,3) (eos) {};
			
			\node[state][io, fill=green, draw=none] at (1.67,0) (y0)                 {}; 
			\node[state][regular polygon, regular polygon sides=3,fill=white,draw=none,minimum width=0.35cm,minimum height=0.35cm,shape border rotate=30,inner sep=0,outer sep=0] at (1.67,0) (sos) {};
			
			\path (y0) edge[att] (s1.south west) (s1) edge[att] (y1) (s2) edge[att] (y2) (s3) edge[att] (y3) (s5) edge[att] (y5);	
			\draw[->,rounded corners] (s1) |- ++(0.5,0.5) -- ++(0,-1) -| (s2.south west);
			\draw[->,rounded corners] (s2) |- ++(0.5,0.5) -- ++(0,-1) -| (s3.south west);
			\draw[->,rounded corners] (s3) |- ++(0.5,0.5) -- ++(0,-1) -| (s4.south west);
			\draw[->,rounded corners] (s4) |- ++(0.5,0.5) -- ++(0,-1) -| (s5.south west);

			
		\end{tikzpicture}\textsl{}
		\caption{An illustration of the length countdown during decoding. The values are represented with learned embeddings or trigonometric encoding.}
		\label{fig:countdown}
	\end{figure}
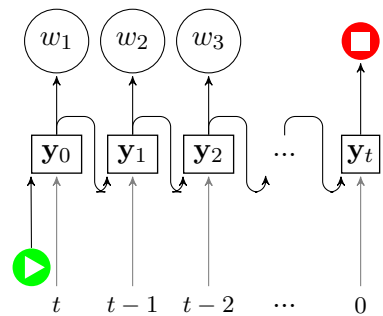

With the baseline introduced above, the ASR model is not aware of the compression task until the adaptation step. 
If the model already has a sense of output length constraints earlier, i.e. when training for the ASR task, it could better utilize the abundance of training data.
Motivated by this hypothesis, we inject information about the allowable output lengths using a count-down at each decoding step, as illustrated in Figure \ref{fig:countdown}.
In a vanilla decoder, the hidden state at position $i$ would ingest an embedding of the previously generated token.
With the count-down for target length $t$, decoder state $\mathbf{y}_i$ additionally ingests a representation of the number of allowed output tokens, $t-i$.

We explore two ways to represent the length count-down.
The first one utilizes length embeddings learned during training, motivated by the approach \citet{kikuchi-etal-2016-controlling} proposed.
Given a target sequence of length $t$, at decoding time step $i$, the input to decoder hidden state $\textbf{y}_i$ is based on a concatenation of the previous state $\textbf{y}_{i-1}$ and an embedding of the remaining length
\begin{equation} \label{eq.lenEmb}
\mathbf{y}_{i-1} \oplus \text{emb}(t-i),
\end{equation}
where emb($t-i$) is an embedding of the number of allowed tokens.
To keep the same dimensionality as that of the original word embedding, the output from Equation \ref{eq.lenEmb} further undergoes a linear transformation followed by the ReLU activation.

With the length embedding approach, the model learns representations of different length values during training. 
Therefore, learning to represent rarely-encountered lengths may be difficult.

The second method modifies the trigonometric positional encoding from the Transformer \cite{vaswani2017attention} to represent the remaining length rather than the current position.
This method has been applied in summarization \cite{takase-okazaki-2019-positional} and machine translation \cite{lakew_surafel_melaku_2019_3524957, niehues2020machine} to limit output lengths.
Motivated by these examples from related sequence generation tasks, we explore the ``backward'' positional encoding in ASR models.

With the original positional encoding, for input dimension $d \in \{0, 1, \dots, D-1\}$, the encoding at position $i$ is defined as
\begin{equation} \label{eq.posEncoding}
\begin{cases}
\text{sin}(i / 10000^{d/D}),& \text{if $d$ is even} \\
\text{cos}(i / 10000^{(d-1)/D}), & \text{if $d$ is odd.} 
\end{cases}
\end{equation}

The backward positional encoding is the same as Equation (\ref{eq.posEncoding}), except that the current position $i$ is replaced by the remaining length $t-i$. Given a target sequence of length $t$, the length encoding at decoding step $i$ becomes 
\begin{equation} \label{eq.lenEnc}
\begin{cases}
\text{sin}((t-i) / 10000^{d/D}),& \text{if $d$ is even} \\
\text{cos}((t-i) / 10000^{(d-1)/D}), & \text{if $d$ is odd.} 
\end{cases}
\end{equation}

Like the positional encoding, the length encoding is also summed together with the input embedding to decoder hidden states.
Moreover, since the encoding is based on sinusoids, it can be easily extrapolated to lengths unseen during training.
This is a potential advantage over the learned length embeddings.

\begin{table*} [h]
	\centering
	\begin{tabular}{lrrrr}
		\toprule
		\textbf{Dataset} & 
		\textbf{Total length (h:m)} & 
		\textbf{Total utterances} &
		\textbf{Average length (s)} & 
		\textbf{Total words}\\
		\midrule
		LibriVoxDeEn (train) & 469:21 & 206,490 & 8.18 & 3,622,560 \\
		LibriVoxDeEn (test) & 5:17 & 2,446 & 7.78 & 51,314 \\
		Tagesschau (adapt) & 37:28 & 11,559  & 11.67 & 243,728 \\
		Tagesschau (test) & 46  & 213 & 13.01  & 4,864 \\
		\bottomrule
	\end{tabular}
	\caption{Corpus statistics.}\label{tab:stats}
\end{table*}

\subsection{Unsupervised Sentence Compression}
Compared to training ASR models to jointly perform transcription and compression, 
a more straightforward approach is to post-edit ASR outputs using a compression model.
However, training such a model in a supervised fashion requires reliable target sequences.
Due to the scarcity of suitable training corpora, we choose an unsupervised approach inspired by multilingual translation \cite{ha2016toward, johnson-etal-2017-googles}. 

Similar to in \citet{niehues2020machine}, this approach relies on a multilingual translation system that is trained on several language pairs.
At training time, language tokens are embedded together with the source and target sentences.
At test time, the model is given the same source and target language token, which is a translation direction unseen in training.
Since the multilingual training enables zero-shot translation, the model is able to reformulate the input in the same language.
To achieve output compression, the length constraints introduced in Section \ref{subsec:e2e.len} are applied in the decoder.

\section{Experiment Setup}
\subsection{Datasets}
Table \ref{tab:stats} provides an overview of audio corpora we use.
The baseline ASR model is trained on the German part of LibriVoxDeEn  \cite{beilharz2019librivoxdeen}, a recently released corpus consisting of open-domain German audio books.
Since the corpus creators did not suggest a train-dev-test partition, we split the dataset ourselves.
The test set contains the following books: \textit{Jonathan Frock}\footnote{In the recordings of \textit{Jonathan Frock}, we found some sections with misaligned utterances therefore incorrect transcriptions. The following sections are excluded from our test set: 00006\_jonathanfrock, 00007\_jonathanfrock.}, 
\textit{Jolanthes Hochzeit} 
and \textit{Kammmacher}.

For the spoken language compression adaptation corpus, 
we collect spoken utterances and subtitles from the German news program Tagesschau from 1 January to 15 August 2019.
To control for recording condition and disfluency, we exclude interviews or press conferences and only keep utterances from the news anchors.
The utterances are segmented based on the start and end time of subtitles.
Since the timestamps do not always precisely correspond to utterance boundaries, we manually verify the test set and edit when necessary.

For the unsupervised compression system, we use the multilingual translation corpus from the IWSLT 2017 evaluation campaign \cite{cettolo2017overview}.
It consists of English, German, Dutch, Italian and Romanian parallel sentences based on TED talks.
All $10\times2$ translation directions are used in training.
At test time, the model is given the same source and target language tag (German in our case) in order to generate summarization in the same language.
We use the positional embedding introduced in Section \ref{subsec:e2e.len} for length control.

\subsection{Preprocessing}
We use the Kaldi toolkit \cite{povey2011kaldi} to preprocess the raw audio utterances into 23-dimensional filter banks.
We choose not to apply any utterance-level normalization to allow for future work towards online processing.
For text materials, i.e. audio transcriptions and the translation source and target sentences, we use byte-pair encoding (BPE)   \cite{sennrich-etal-2016-neural} to create subword-based dictionaries.


\subsection{Hyperparameters}
For the ASR model, we adopt many reported values in the work of \citet{pham2019very}, including the optimizer choice, learning rate, warmup steps, dropout rate, label smoothing rate, and embedding dimension.
There are several parameters that we choose differently.
The size of the inner feed forward layer is 2048. 
Moreover, we use 32 encoder and 12 decoder layers, and BPE of size 10,000.
For the compression model, we use a Transformer with 8 encoder and decoder layers each.\footnote{The code is available at \url{https://github.com/quanpn90/NMTGMinor/tree/DbMajor}.}

\section{Experiments}
\subsection{Post-Editing with Compression Model}
To gain an initial understanding of the task, we start with a more controlled setup, where the test utterances are transcribed by a commercial off-the-shelf ASR system.
The transcriptions are then post-processed with our compression model.

\begin{figure}[h]
	\centering
	\includegraphics[width=\columnwidth]{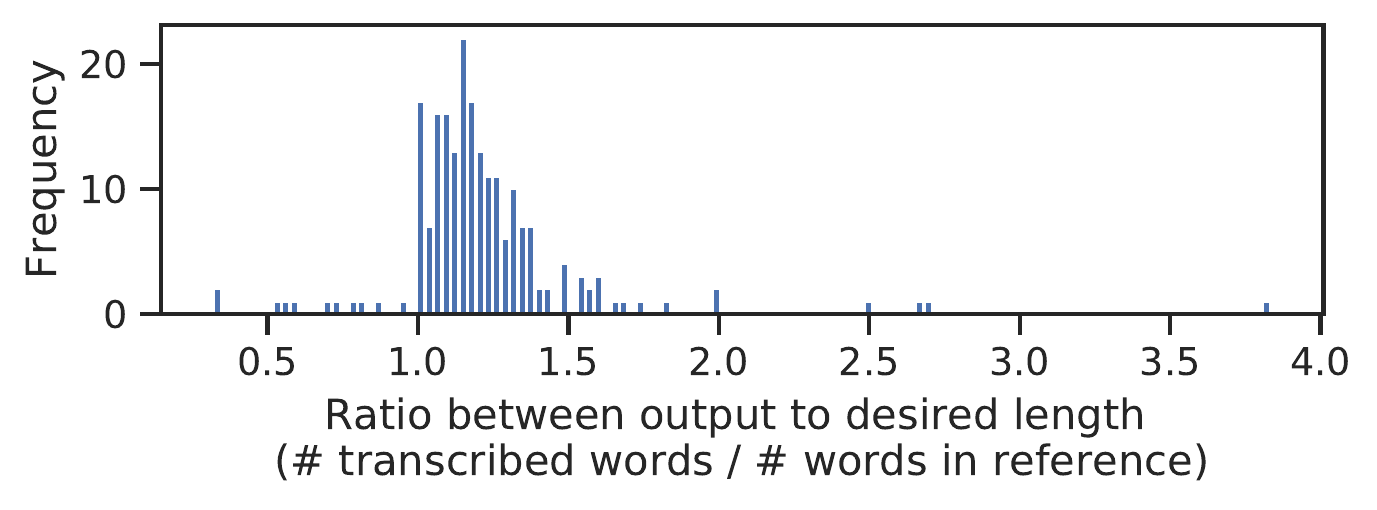}
	\caption{Histogram of compression rate on the test set. Ratios lower than 1 are outliers where the off-the-shelf ASR model terminates decoding prematurely.}
	\label{fig:hist}
\end{figure}

\begin{table}[h]
	\centering
	\small 
	\begin{tabularx}{\columnwidth}{X}
		\toprule
		\textbf{Ground-truth}: (17 words) Allerdings k{\"o}nnen Ausnahmeantr{\"a}ge gestellt werden, von Pendlern und von Stuttgartern, was Tausende auch bereits getan haben.  {\color{gray} (However, commuters and Stuttgarter can apply for exception-permits, which thousands have already done.) }\\
		\textbf{Reference}: (6 words) Es k{\"o}nnen aber Ausnahmeanträge gestellt werden.  
		{\color{gray} (However, exception-permits can be applied for.)} \\
		\midrule
		\textbf{Ground-truth}: (13 words) In der Debatte {\"u}ber neue Regeln f{\"u}r Organspenden gibt es einen ersten Gesetzentwurf.  
		{\color{gray} (There is a first draft law in the debate about new rules for organ donation.)}\\ 
		\textbf{Reference}: (12 words) In der Debatte {\"u}ber neue Regeln f{\"u}r Organspenden gibt es einen Gesetzentwurf. {\color{gray} (There is a draft law in the debate about new rules for organ donation.)} \\
		\bottomrule
	\end{tabularx}	
	\caption{Two examples of various levels of compression, where the first shortens from 17 to 6 words, and the second only removes one word. }\label{tab:example.shorten}
\end{table}


First, we analyze the level of desired output compression by contrasting the lengths of the off-the-shelf transcriptions against those of the references.
In Figure \ref{fig:hist}, we plot the distribution of the ratio between transcription lengths and target lengths over the test set.
The first observation is that most of the transcriptions require shortening, as shown by the high frequencies of ratios over 1.

\begin{table*}[h]
	\centering
	\begin{tabular}{lccccc}
		\toprule
		\textbf{Model} & \textbf{Ratio (output to desired length)} & \textbf{WER} & \textbf{R-1} & \textbf{R-2} & \textbf{R-L} \\
		\midrule
		Off-the-shelf ASR & 1.09 & 41.5  & 75.9    & 58.9  & 72.6 \\ 
		$+$ compression   & 0.95 & 69.1  & 55.7    & 29.5  & 51.6 \\ 
		\midrule
		Baseline ASR     & 1.21 & 57.7  & 65.2    & 43.6  & 61.9 \\ 
		$+$ compression  & 1.00 & 74.3  & 48.5    & 23.3  & 44.7 \\ 
		\bottomrule
	\end{tabular}
	\caption{Word error rate and summarization quality on the test set by cascading two separate models for ASR and sentence compression. }\label{tab:1}
\end{table*}

\begin{table*}[h]
	\centering
	\begin{tabularx}{\textwidth}{p{2.3cm}X}
		\toprule
		\textbf{Ground-truth} 
		&
		Es ist kurz nach Mitternacht, als pl{\"o}tzlich ein Auto in eine Gruppe von Menschen steuert, die ausgelassen ins neue Jahr feiern. 
		{\color{gray}It is just after midnight, when a car suddenly drives into a group of people who joyfully celebrate the new year.}\\
		\textbf{Reference} 
		& Kurz nach Mitternacht steuert ein Auto in eine Gruppe von Menschen, die ins neue Jahr feiern. 
		{\color{gray} Just after midnight a car drives into a group of people who celebrate the new year.} \\
		\textbf{Unsupervised \newline  compression} 
		& Kurz nach Mitternacht \underline{f{\"a}hrt} ein Auto pl{\"o}tzlich in eine Gruppe von \underline{Leuten}, die das \underline{n{\"a}chste} Jahr feiern. 
		{\color{gray} Just after midnight a car suddenly drives into a group of people who celebrate the \underline{next} year.}\\
		\midrule
		\textbf{Ground-truth} 
		& Unter dem Eindruck der Massenproteste hatten sich zuletzt auch hochrangige Milit{\"a}rs von ihm abgewandt. 
		{\color{gray} Under the impression of mass protests, senior military officials have finally also turned away from him.} \\
		\textbf{Reference} 
		& Unter dem Eindruck der Proteste wandten sich zuletzt auch hochrangige Milit{\"a}rs ab. 
		{\color{gray} Under the impression of protests, senior military officials finally also turned away.} \\
		\textbf{Unsupervised \newline  compression} 
		& Unter dem Eindruck \underline{von} Massenprotest\underline{ieren} \underline{waren} auch hochrangige Milit{\"a}rs von ihm \underline{entfernt}. 
		{\color{gray}Under the impression of mass protest\underline{ing}, senior military officials \underline{were} also \underline{distanced} from him.} \\
		\bottomrule
	\end{tabularx}
	\caption{
		\label{tab:examples} Examples outputs of the unsupervised compression model, where undesired paraphrasing is underlined. English translations are in gray.
	}
\end{table*}

Moreover, the compression ratio varies across different utterances.
Table \ref{tab:example.shorten} shows two examples, where the first compressed from 17 to 6 words, while the second only deletes  one word. 
When inspecting the original videos, we notice that the first example contains many other visual contents, some also in text form, whereas the second example only involves the new anchorwoman speaking.
The examples showcase that the level of desired compression depends on various factors, such as the amount of visual information simultaneously shown on the screen.
Therefore, a globally fixed compression rate would not be suitable.

To comply with length constraints, the ASR outputs are shortened by the unsupervised sentence compression model.
As the system is trained based on subwords, we use the number of BPE-tokens in the reference as target length.
The first two rows in Table \ref{tab:1} contrast the output quality before and after compression, as measured in case-insensitive word error rate (WER), and ROUGE scores \cite{lin2004rouge}.\footnote{Ideally, these metrics should be accompanied by an independent human evaluation, which we could not perform due to resource constraints.}
To our surprise, compression has a large negative impact on the outputs in all four metrics, creating a gap of over 20\% absolute.
Via an exhaustive manual inspection over the test set, we find that the unsupervised compression model tends to paraphrase much more frequently than the references.
While the paraphrased output is often valid both grammatically and semantically, the deviation from the references leads to higher WER and lower ROUGE scores.
Two examples are given in Table \ref{tab:examples}, where several synonym replacements appear in the compression outputs, e.g. ``f{\"a}hrt'' for ``steuert'' (both ``drives''), ``Leute'' for ``Menschen'' (both ``people''), ``n{\"a}chste Jahr'' for ``neue Jahr'' (``next year'' for ``new year'').
In all these places, the references keep the original spoken words unchanged.
Given the nature of our task, it is indeed undesirable to paraphrase excessively, as subtitles that are too different from the original spoken utterances could create a cognitive overload to users.

\begin{table*}[h]
	\centering
	\begin{tabular}{lccccc}
		\toprule
		\textbf{Model} & \textbf{Ratio (output to desired length)} & \textbf{WER} & \textbf{R-1} & \textbf{R-2} & \textbf{R-L} \\
		\midrule
		Baseline ASR & 1.12 & 57.7  & 65.2    & 43.6  & 61.9 \\
		$+$adapt   & 1.05 & 38.5  & 76.8    & 58.7  & 74.4 \\
		$+$adapt$+$stop decoding   & 0.96 & 39.9  & 74.6    & 57.0  & 72.6\\
		\bottomrule
	\end{tabular}
	\caption{Word error rate and summarization quality on the test set by the baseline ASR model. Fine-tuning on in-domain data is beneficial for both shortening outputs and increasing transcription quality. }\label{tab:2}
\end{table*}

\begin{table*}[h]
	\centering
	\begin{tabularx}{\textwidth}{p{3cm}X}
		\toprule
		\textbf{Reference}
		&
		In Brasilien ist Pr{\"a}sident Bolsonaro vereidigt worden. 
		\newline
		{\color{gray} In Brazil new president Bolsonaro has been inaugurated. }\\
		\textbf{Before adaptation}
		&
		In Brasilien \underline{ist} der neue Pr{\"a}sident \textit{voll zu Narro} \underline{vereidigt worden}. \newline
		{\color{gray} In Brazil the new president ``\textit{voll zu Narro}'' has been inaugurated. }\\
		\textbf{After  adaptation}
		& In Brasilien \underline{wurde} der neue Pr{\"a}sident \textit{Bolsonaro} \underline{vereidigt}. \newline
		{\color{gray} In Brazil the new president \textit{Bolsonaro} was inaugurated. }\\
		\bottomrule
	\end{tabularx}
	\caption{
		\label{tab:examples2} Example outputs of the ASR model before and after adapting to the compression task. After adaptation, the model can perform shortening by changing tense, and correctly recognize the proper noun ``Bolsonaro''. English translations are below each sentence in gray.
	}
\end{table*}

Considering these downsides, an ASR system is trained from scratch to provide more flexibility of structural modification.
On our self-partitioned LibriVoxDeEn test set, we achieve a WER of 9.2\%.
The compression performance is reported in the lower section of Table \ref{tab:1}.
As this model is only trained on LibriVoxDeEn audio books,
its performance suffers from the train-test domain mismatch.
This is exhibited by the gap of nearly 10\% to the off-the-shelf system, which is trained on larger volume of data from various domains.
Moreover, the transcription errors by the ASR system carry over as the input of the compression model, which is further disadvantageous to the final output quality.
Lastly, similar to previous observations, post-processing by the compression model has a negative effect in terms of the evaluation metrics. 

Overall, the performance of the unsupervised compression model suffers from paraphrasing.
As an anonymous reviewer suggested, the aggressive paraphrasing can be remedied during decoding. 
For example, given a large beam size, we can select candidates that contain less paraphrasing. 
Otherwise, training with a paraphrasing penalty could also alleviate the problem.
While we have not explored these methods here, they would indeed provide a more complete picture when comparing the cascaded and end-to-end approach.

\subsection{Fine-Tuning ASR Model for Compression}
Our baseline ASR model is trained on a different domain than the test set, and only for the transcription task. 
To improve performance on our task of interest, we apply fine-tuning on the adaptation corpus.
The results are in Table \ref{tab:2}.
For easy visual comparison, the pre-adaptation performance in the first row is repeated from Table \ref{tab:1}.
Contrasting the performance before and after adaptation,
we see noticeable gains brought by adaptation in terms of all four quality metrics.
Moreover, as evidenced by the reduced ratio of output to desired length, the model already performs shortening, despite not having received explicit length constraints.
Table \ref{tab:examples2} shows an example where the adapted model changes verb tense to reduce output length.
Specifically, by using the simple past (``wurde vereidigt / was inaugurated'') instead of the present perfect tense (``ist vereidigt worden / has been inaugurated''), the output becomes shorter.
Meanwhile, we also observe the correct transcription of the proper noun ``Bolsonaro'', which was mistakenly transcribed to the phonetically-similar ``voll zu Narro'' before adaption.
This illustrates that the adaptation step enables the model to improve recognition quality and compress its outputs simultaneously.

\begin{table*}[h]
	\centering
	\begin{tabular}{lccccc}
		\toprule
		\textbf{Model} & \textbf{Ratio (output to desired length)} & \textbf{WER} & \textbf{R-1} & \textbf{R-2} & \textbf{R-L} \\
		\midrule
		(1): Baseline (satisfy length)    & 0.97 & 55.1  & 62.5    & 41.5
		& 60.4 \\ 
		(2): \qquad \qquad \qquad \qquad $+$ adapt  & 0.96 & 39.9  & 74.6    & \textbf{57.0}  & 72.6 \\
		\midrule
		(3): Length embedding &  1.00  & 57.8  &   61.9   & 40.5  & 59.8  \\
		(4): \qquad \qquad \qquad \qquad $+$ adapt & 0.96 & 39.3  & 74.3    & 55.2  & 72.5 \\
		\midrule
		(5): Length encoding &  1.00  & 57.4  &   62.6  & 40.7  & 60.1 \\
		(6): \qquad \qquad \qquad \qquad $+$ adapt & 0.96 & \textbf{38.6}  & \textbf{75.1}   & 56.4  & \textbf{73.2} \\
		\bottomrule
	\end{tabular}
	\caption{Word error rate and summarization quality by the models with length count-down. Adaptation results in large gains. The model with length encoding outperforms the baseline  and the one with length embedding.}\label{tab:3}
\end{table*}
\begin{table*}
	\centering
	\begin{tabularx}{\textwidth}{p{3cm}X}
		\toprule
		\textbf{Reference}
		&
		[... 75 tokens]  
		Sea-Watch spricht von einer politisch motivierten Blockade, um Rettungsaktionen zu verhindern. {\color{gray} Sea-Watch speaks of a politically motivated blockade to prevent bailouts.}\\
		\textbf{Len. Encoding}
		& [... 83 tokens] Die Hilfsorganisation Sea-Watch spricht von einer politisch motivierten Blockade. {\color{gray} The aid organization Sea-Watch speaks of a politically motivated blockade.}\\
		\textbf{Len. Embedding}
		& [... 82 tokens] Die Hilfsorganisation Sea-Watch spricht von einer politisch motivierten Blockade zu verhinder. {\color{gray} The aid organization Sea-Watch speaks of a politically motivated blockade to prevent. }\\
		\bottomrule
	\end{tabularx}
	\caption{
		\label{tab:examples3} Comparison of length encoding and length embedding models (after adaptation). When facing exceptionally long outputs, the length embedding model tends to stop abruptly, producing non-grammatical half-sentences.
	}
\end{table*}

Despite the positive observations, the desired length constraints are not yet fully satisfied, as shown by the ratio of 1.05 between output to desired lengths.
To examine the scenario of fully obeying the length constraints, we stop decoding once the number of allowed tokens runs out. 
The result is reported in the last row of Table \ref{tab:2}.
The ratio of 0.96 is lower than 1 because of a few instances where decoding stops before the count-down reaches zero.
As the same output sequence can be constructed by different BPE-units, choosing longer subword units earlier on can lead to reaching the end-of-sequence token before depleting the number of allowed tokens.
Lastly, from the quality metrics in the last row of Table \ref{tab:2}, we see that the forced termination of decoding comes with higher WER and lower ROUGE scores, indicating reduced output quality when fully satisfying the target length constraints.

\subsection{Models with Explicit Length Constraints}
While the baseline ASR model achieves some degree of compression after adaptation, it cannot fully comply with length constraints.
Therefore, the following experiments examine the effects of training with explicit length count-downs.
In Table \ref{tab:3}, we report the performance of the ASR models with length embedding or encoding, as introduced in Section \ref{subsec:e2e.len}.
For a complete comparison, in the first two rows of Table \ref{tab:3}, we also include the baseline performance with forced termination of decoding.

Rows (3) and (5) show the performance of the two length count-down methods before adaptation.
As the ratios of output to desired length are equal to 1, the models are always faithful to the given length constraints.
This shows the effectiveness of injecting allowable length in training.
However, we also observe that there is no quality gain over the baseline in row (1).
To investigate the reason for this, we experimented by decoding one sample sequence with different allowable lengths.
As we gradually reduce the target length, the models first shorten their outputs by removing punctuation marks.
Afterwards, instead of shortening the outputs by summarization, they stop decoding when running out of allowed number of tokens.
Indeed, during training, the models are only incentivized to accurately transcribe the spoken utterance and to stop decoding when the count-down reaches zero.
The same behavior therefore carries over to test time.
Contrary to abstractive summarization \cite{takase-okazaki-2019-positional} and machine translation \cite{lakew_surafel_melaku_2019_3524957}, in ASR, an input sequence has one single ground-truth transcription rather than multiple viable outputs.
This could lead to a different level of abstraction than required in summarization or translation models.
These observations with the unadapted model also highlight the importance of the subsequent fine-tuning step.

The results after adaptation are reported in rows (4) and (6). 
The large improvement in WER and ROUGE scores is in line with the previous finding when adapting the baseline.
When decoding with length count-down, we define the target output length as the minimum of the baseline output length and the reference length.
This ensures the same output-to-desired length ratio as the baseline in row (1).
Here, the first observation is that the length encoding model in row (5) outperforms the baseline in three of the four evaluation metrics.
This suggests that it is beneficial to represent the constraints explicitly during training.
Moreover, the length encoding model also consistently outperforms its embedding-based counterpart.
This could be because the length encoding can extrapolate to any length values, and 
is equipped with a sense of relative
differences between numerical values at initialization. 
On the other hand, the length embedding would need to learn the representation for different lengths during training.
When inspecting the outputs for long utterances, we found that the embedding model is more likely to abruptly stop, such as the example shown in Table \ref{tab:examples3}.

\section{Related Work}
\subsection{Length-Controlled Text Generation}
Controlling output length of natural language generation systems has been studied for several tasks.

For abstractive summarization, \citet{kikuchi-etal-2016-controlling} proposed two methods to incorporate length constraints into LSTM-based encoder-decoder models. The first method uses length embedding at every decoding step, while the second adds the desired length in the first decoder state.
For convolutional models, \citet{fan-etal-2018-controllable} used special tokens to represent quantized length ranges, and provides the desired token to the decoder before output generation.
\citet{liu-etal-2018-controlling} adopted a more general approach, where the decoder directly ingests the desired length.
More recently, \citet{takase-okazaki-2019-positional} modified the positional encoding from the Transformer \cite{vaswani2017attention} to encode allowable lengths.
\citet{makino-etal-2019-global} proposed a loss function that encourages summaries within desired lengths. 
\citet{saito2020length} introduced a model that controls both output length and informativeness.

For machine translation, \citet{lakew_surafel_melaku_2019_3524957} used both the length range token and reverse length encoding.
\citet{niehues2020machine} used the length embedding, encoding, as well as a combination of the original positional encoding and length count-down.
\subsection{Sentence Compression}
Our task of length-controlled ASR outputs is related to sentence compression, as the transcriptions can be compressed in post-processing.
An early approach of supervised extractive sentence compression was by \citet{filippova-etal-2015-sentence}, who proposed to predict the delete-or-keep choice for each output symbol.
\citet{angerbauer2019automatic} extended this approach by integrating the desired compression ratio as part of the prediction label.
\citet{yu-etal-2018-operation} proposed to combine the merits of extractive and abstractive approaches by first deleting on non-essential words and then generating new words.
For unsupervised compression, \citet{fevry-phang-2018-unsupervised} trained a denoising auto-encoder to reconstruct original sentences, and in this way circumvented the need for supervised corpora.

\section{Conclusion}
In this work, we explored the task of compressing ASR outputs to enhance subtitle readability.
This task has several unique properties. 
First, the compression is not solely deletion-based.
Moreover, unnecessary paraphrasing must be limited to maintain a consistent user experience between hearing and reading.

We first investigated cascading an ASR module with a sentence compression model.
Due to the absence of supervised corpora, the compression model is trained in an unsupervised fashion.
Experiments showed that the outputs generated this way do not suit our task requirements because of unnecessary paraphrasing.
We then adapted an end-to-end ASR model on a small corpus with compressed transcriptions. 
Via adaptation, the model learned to both shorten its outputs and improve transcription quality.
Nevertheless, the given length constraints were not fully satisfied.
Lastly, by explicitly injecting length constraints via reverse positional encoding, we achieved further performance gain, while completely adhering to length constraints.

A direction for future work is to incorporate more diverse measurements of output length as well as complexity.
In this work, we measured length by the number of BPE-tokens.
While this typically corresponds to the output length perceived visually, a more direct metric would be the number of characters.
Moreover, output complexity, such as the proportion of long words, is also important for readability and therefore worth exploring.
In a broader scope, as an anonymous reviewer suggested, a way to alleviate the resource scarcity for end-to-end ASR compression is to augment the training data with synthesized utterances from summarization corpora.
We expect the augmentation to be complementary to our approaches in this work. 

\bibliography{anthology,acl2020}

\begin{thebibliography}{28}
\expandafter\ifx\csname natexlab\endcsname\relax\def\natexlab#1{#1}\fi

\bibitem[{Angerbauer et~al.(2019)Angerbauer, Adel, and
  Vu}]{angerbauer2019automatic}
Katrin Angerbauer, Heike Adel, and Ngoc~Thang Vu. 2019.
\newblock \href {https://doi.org/10.21437/Interspeech.2019-1750} {Automatic
  compression of subtitles with neural networks and its effect on user
  experience}.
\newblock In \emph{Proceedings of Interspeech'2019}, pages 594--598.

\bibitem[{Beilharz et~al.(2020)Beilharz, Sun, Karimova, and
  Riezler}]{beilharz2019librivoxdeen}
Benjamin Beilharz, Xin Sun, Sariya Karimova, and Stefan Riezler. 2020.
\newblock \href {https://arxiv.org/pdf/1910.07924.pdf} {{LibriVoxDeEn}: A
  corpus for {G}erman-to-{E}nglish speech translation and speech recognition}.
\newblock In \emph{Proceedings of LREC'2020}.

\bibitem[{Cettolo et~al.(2017)Cettolo, Federico, Bentivogli, Niehues,
  St{\"u}ker, Sudoh, Yoshino, and Federmann}]{cettolo2017overview}
Mauro Cettolo, Marcello Federico, Luisa Bentivogli, Jan Niehues, Sebastian
  St{\"u}ker, Katsuitho Sudoh, Koichiro Yoshino, and Christian Federmann. 2017.
\newblock \href {http://workshop2017.iwslt.org/downloads/Report-Paper.pdf}
  {Overview of the {IWSLT} 2017 evaluation campaign}.
\newblock In \emph{Proceedings of IWSLT'2017}, pages 2--14.

\bibitem[{Chopra et~al.(2016)Chopra, Auli, and
  Rush}]{chopra-etal-2016-abstractive}
Sumit Chopra, Michael Auli, and Alexander~M. Rush. 2016.
\newblock \href {https://doi.org/10.18653/v1/N16-1012} {Abstractive sentence
  summarization with attentive recurrent neural networks}.
\newblock In \emph{Proceedings of NAACL-HLT'2016}, pages 93--98.

\bibitem[{Clarke and Lapata(2006)}]{clarke-lapata-2006-models}
James Clarke and Mirella Lapata. 2006.
\newblock \href {https://doi.org/10.3115/1220175.1220223} {Models for sentence
  compression: A comparison across domains, training requirements and
  evaluation measures}.
\newblock In \emph{Proceedings of COLING/ACL'2006}, pages 377--384.

\bibitem[{Cohn and Lapata(2008)}]{cohn-lapata-2008-sentence}
Trevor Cohn and Mirella Lapata. 2008.
\newblock \href {https://www.aclweb.org/anthology/C08-1018} {Sentence
  compression beyond word deletion}.
\newblock In \emph{Proceedings of COLING'2008}, pages 137--144.

\bibitem[{Fan et~al.(2018)Fan, Grangier, and Auli}]{fan-etal-2018-controllable}
Angela Fan, David Grangier, and Michael Auli. 2018.
\newblock \href {https://doi.org/10.18653/v1/W18-2706} {Controllable
  abstractive summarization}.
\newblock In \emph{Proceedings of WNMT'2018}, pages 45--54, Melbourne,
  Australia. Association for Computational Linguistics.

\bibitem[{Fevry and Phang(2018)}]{fevry-phang-2018-unsupervised}
Thibault Fevry and Jason Phang. 2018.
\newblock \href {https://doi.org/10.18653/v1/K18-1040} {Unsupervised sentence
  compression using denoising auto-encoders}.
\newblock In \emph{Proceedings of CoNLL'2018}, pages 413--422.

\bibitem[{Filippova et~al.(2015)Filippova, Alfonseca, Colmenares, Kaiser, and
  Vinyals}]{filippova-etal-2015-sentence}
Katja Filippova, Enrique Alfonseca, Carlos~A. Colmenares, Lukasz Kaiser, and
  Oriol Vinyals. 2015.
\newblock \href {https://doi.org/10.18653/v1/D15-1042} {Sentence compression by
  deletion with {LSTM}s}.
\newblock In \emph{Proceedings of EMNLP'2015}, pages 360--368.

\bibitem[{Ha et~al.(2017)Ha, Niehues, and Waibel}]{ha2016toward}
Thanh-Le Ha, Jan Niehues, and Alexander Waibel. 2017.
\newblock \href
  {https://workshop2016.iwslt.org/downloads/IWSLT_2016_paper_5.pdf} {Toward
  multilingual neural machine translation with universal encoder and decoder}.
\newblock In \emph{Proceedings of IWSLT'2016}.

\bibitem[{{He} et~al.(2019){He}, {Sainath}, {Prabhavalkar}, {McGraw},
  {Alvarez}, {Zhao}, {Rybach}, {Kannan}, {Wu}, {Pang}, {Liang}, {Bhatia},
  {Shangguan}, {Li}, {Pundak}, {Sim}, {Bagby}, {Chang}, {Rao}, and
  {Gruenstein}}]{he2019streaming}
Y.~{He}, T.~N. {Sainath}, R.~{Prabhavalkar}, I.~{McGraw}, R.~{Alvarez},
  D.~{Zhao}, D.~{Rybach}, A.~{Kannan}, Y.~{Wu}, R.~{Pang}, Q.~{Liang},
  D.~{Bhatia}, Y.~{Shangguan}, B.~{Li}, G.~{Pundak}, K.~C. {Sim}, T.~{Bagby},
  S.~{Chang}, K.~{Rao}, and A.~{Gruenstein}. 2019.
\newblock \href {https://doi.org/10.1109/ICASSP.2019.8682336} {Streaming
  end-to-end speech recognition for mobile devices}.
\newblock In \emph{Proceedings of ICASSP'2019}, pages 6381--6385.

\bibitem[{Johnson et~al.(2017)Johnson, Schuster, Le, Krikun, Wu, Chen, Thorat,
  Vi{\'e}gas, Wattenberg, Corrado, Hughes, and
  Dean}]{johnson-etal-2017-googles}
Melvin Johnson, Mike Schuster, Quoc~V. Le, Maxim Krikun, Yonghui Wu, Zhifeng
  Chen, Nikhil Thorat, Fernanda Vi{\'e}gas, Martin Wattenberg, Greg Corrado,
  Macduff Hughes, and Jeffrey Dean. 2017.
\newblock \href {https://doi.org/10.1162/tacl_a_00065} {{G}oogle{'}s
  multilingual neural machine translation system: Enabling zero-shot
  translation}.
\newblock \emph{Transactions of the Association for Computational Linguistics},
  5:339--351.

\bibitem[{Kikuchi et~al.(2016)Kikuchi, Neubig, Sasano, Takamura, and
  Okumura}]{kikuchi-etal-2016-controlling}
Yuta Kikuchi, Graham Neubig, Ryohei Sasano, Hiroya Takamura, and Manabu
  Okumura. 2016.
\newblock \href {https://doi.org/10.18653/v1/D16-1140} {Controlling output
  length in neural encoder-decoders}.
\newblock In \emph{Proceedings of EMNLP'2016}, pages 1328--1338.

\bibitem[{Knight and Marcu(2002)}]{knight2002summarization}
Kevin Knight and Daniel Marcu. 2002.
\newblock Summarization beyond sentence extraction: A probabilistic approach to
  sentence compression.
\newblock \emph{Artificial Intelligence}, 139(1):91--107.

\bibitem[{Lakew et~al.(2019)Lakew, Gangi, and
  Federico}]{lakew_surafel_melaku_2019_3524957}
Surafel~Melaku Lakew, Mattia~Di Gangi, and Marcello Federico. 2019.
\newblock \href {https://doi.org/10.5281/zenodo.3524957} {Controlling the
  output length of neural machine translation}.
\newblock In \emph{Proceedings of IWSLT'2019}.

\bibitem[{Li et~al.(2017)Li, Sainath, Narayanan, Caroselli, Bacchiani, Misra,
  Shafran, Sak, Pundak, Chin et~al.}]{li2017acoustic}
Bo~Li, Tara~N Sainath, Arun Narayanan, Joe Caroselli, Michiel Bacchiani, Ananya
  Misra, Izhak Shafran, Hasim Sak, Golan Pundak, Kean~K Chin, et~al. 2017.
\newblock Acoustic modeling for google home.
\newblock In \emph{Proceedings of Interspeech'2017}, pages 399--403.

\bibitem[{Lin(2004)}]{lin2004rouge}
{Chin-Yew} Lin. 2004.
\newblock \href {https://www.aclweb.org/anthology/W04-1013} {{ROUGE}: A package
  for automatic evaluation of summaries}.
\newblock In \emph{{Text Summarization Branches Out}}, pages 74--81, Barcelona,
  Spain. Association for Computational Linguistics.

\bibitem[{Liu et~al.(2018)Liu, Luo, and Zhu}]{liu-etal-2018-controlling}
Yizhu Liu, Zhiyi Luo, and Kenny Zhu. 2018.
\newblock \href {https://doi.org/10.18653/v1/D18-1444} {Controlling length in
  abstractive summarization using a convolutional neural network}.
\newblock In \emph{Proceedings of EMNLP'2018}, pages 4110--4119.

\bibitem[{Makino et~al.(2019)Makino, Iwakura, Takamura, and
  Okumura}]{makino-etal-2019-global}
Takuya Makino, Tomoya Iwakura, Hiroya Takamura, and Manabu Okumura. 2019.
\newblock \href {https://doi.org/10.18653/v1/P19-1099} {Global optimization
  under length constraint for neural text summarization}.
\newblock In \emph{Proceedings of ACL'2019}, pages 1039--1048.

\bibitem[{Niehues(2020)}]{niehues2020machine}
Jan Niehues. 2020.
\newblock \href {http://arxiv.org/abs/2004.03176} {Machine translation with
  unsupervised length-constraints}.
\newblock \emph{arXiv preprint arXiv:2004.03176}.

\bibitem[{Pham et~al.(2019)Pham, Nguyen, Niehues, M{\"u}ller, and
  Waibel}]{pham2019very}
Ngoc{-}Quan Pham, Thai{-}Son Nguyen, Jan Niehues, Markus M{\"u}ller, and Alex
  Waibel. 2019.
\newblock \href {https://doi.org/10.21437/Interspeech.2019-2702} {Very deep
  self-attention networks for end-to-end speech recognition}.
\newblock In \emph{Proceedings of Interspeech'2019}, pages 66--70.

\bibitem[{Povey et~al.(2011)Povey, Ghoshal, Boulianne, Burget, Glembek, Goel,
  Hannemann, Motlicek, Qian, Schwarz, Silovsky, Stemmer, and
  Vesely}]{povey2011kaldi}
Daniel Povey, Arnab Ghoshal, Gilles Boulianne, Lukas Burget, Ondrej Glembek,
  Nagendra Goel, Mirko Hannemann, Petr Motlicek, Yanmin Qian, Petr Schwarz, Jan
  Silovsky, Georg Stemmer, and Karel Vesely. 2011.
\newblock The {K}aldi speech recognition toolkit.
\newblock In \emph{Proceedings of ASRU'2011}.

\bibitem[{Rush et~al.(2015)Rush, Chopra, and Weston}]{rush-etal-2015-neural}
Alexander~M. Rush, Sumit Chopra, and Jason Weston. 2015.
\newblock \href {https://doi.org/10.18653/v1/D15-1044} {A neural attention
  model for abstractive sentence summarization}.
\newblock In \emph{Proceedings EMNLP'2015}, pages 379--389.

\bibitem[{Saito et~al.(2020)Saito, Nishida, Nishida, Otsuka, Asano, Tomita,
  Shindo, and Matsumoto}]{saito2020length}
Itsumi Saito, Kyosuke Nishida, Kosuke Nishida, Atsushi Otsuka, Hisako Asano,
  Junji Tomita, Hiroyuki Shindo, and Yuji Matsumoto. 2020.
\newblock \href {https://arxiv.org/abs/2001.07331} {Length-controllable
  abstractive summarization by guiding with summary prototype}.
\newblock \emph{arXiv preprint arXiv:2001.07331}.

\bibitem[{Sennrich et~al.(2016)Sennrich, Haddow, and
  Birch}]{sennrich-etal-2016-neural}
Rico Sennrich, Barry Haddow, and Alexandra Birch. 2016.
\newblock \href {https://doi.org/10.18653/v1/P16-1162} {Neural machine
  translation of rare words with subword units}.
\newblock In \emph{Proceedings of ACL'2016}, pages 1715--1725.

\bibitem[{Takase and Okazaki(2019)}]{takase-okazaki-2019-positional}
Sho Takase and Naoaki Okazaki. 2019.
\newblock \href {https://doi.org/10.18653/v1/N19-1401} {Positional encoding to
  control output sequence length}.
\newblock In \emph{Proceedings NAACL-HLT'2019}, pages 3999--4004.

\bibitem[{Vaswani et~al.(2017)Vaswani, Shazeer, Parmar, Uszkoreit, Jones,
  Gomez, Kaiser, and Polosukhin}]{vaswani2017attention}
Ashish Vaswani, Noam Shazeer, Niki Parmar, Jakob Uszkoreit, Llion Jones,
  Aidan~N. Gomez, undefinedukasz Kaiser, and Illia Polosukhin. 2017.
\newblock Attention is all you need.
\newblock In \emph{Proceedings of NIPS'2017}, pages 6000--6010.

\bibitem[{Yu et~al.(2018)Yu, Zhang, Huang, and Zhu}]{yu-etal-2018-operation}
Naitong Yu, Jie Zhang, Minlie Huang, and Xiaoyan Zhu. 2018.
\newblock \href {https://www.aclweb.org/anthology/C18-1091} {An operation
  network for abstractive sentence compression}.
\newblock In \emph{Proceedings of COLING'2018}, pages 1065--1076.

\end{thebibliography}
\bibliographystyle{acl_natbib}

\clearpage
\end{document}